\pdfoutput=1

\documentclass[11pt]{article}

\usepackage[]{acl}

\usepackage{times}
\usepackage{latexsym}

\usepackage[T1]{fontenc}

\usepackage[utf8]{inputenc}

\usepackage{microtype}

\usepackage{inconsolata}
\usepackage{inconsolata}
\usepackage{graphicx}
\usepackage{stfloats}
\usepackage{multirow}
\usepackage{subfigure} 
\usepackage{makecell}
\usepackage{caption}
\usepackage{subfigure}
\usepackage{arydshln}
\usepackage{booktabs}
\usepackage{multirow}
\usepackage{tablefootnote}
\usepackage{algorithm}
\usepackage{algorithmic}
\usepackage{amsmath}
\usepackage{multirow}
\usepackage{arydshln}
\usepackage{booktabs}
\usepackage{newfloat}
\usepackage{listings}
\usepackage{soul}
\usepackage{float}
\usepackage{amsfonts}

\usepackage[]{todonotes}

\newcommand{\frameworkname}{\textsc{InFO }}
\newcommand{\frameworknamens}{\textsc{InFO}}

\newcommand{\modelname}{{\frameworknamens}-\textsc{RAG }}
\newcommand{\modelnamens}{{\frameworknamens}-\textsc{RAG}}

\title{Unsupervised Information Refinement Training of Large Language Models for Retrieval-Augmented Generation}

\author{%
Shicheng Xu$^{1,2}$\thanks{\,\,Work done during the Tencent Rhino-bird Research Elite Program at WeChat.} \quad Liang Pang$^{1}$\thanks{\ \ Corresponding authors.} \quad Mo Yu$^{3}$\footnotemark[2]
\quad  Fandong Meng$^{3}$ \quad Huawei Shen$^{1}$ \\ \bf  \quad Xueqi Cheng$^{1}$ \quad Jie Zhou$^{3}$ \\
$^{1}$CAS Key Laboratory of AI Safety,
 Institute of Computing Technology, CAS \\
 $^{2}$University of Chinese Academy of Sciences\\
 $^{3}$Pattern Recognition Center, WeChat AI \\
  {\small{\texttt{\{xushicheng21s,pangliang,shenhuawei,cxq\}@ict.ac.cn}}} \\
  \small{\texttt{{moyumyu@global.tencent.com} \quad \{fandongmeng,withtomzhou\}@tencent.com}}}
  
\begin{document}
\maketitle
\begin{abstract}

Retrieval-augmented generation (RAG) enhances large language models (LLMs) by incorporating additional information from retrieval. However, studies have shown that LLMs still face challenges in effectively using the retrieved information, even ignoring it or being misled by it. The key reason is that the training of LLMs does not clearly make LLMs learn how to utilize input retrieved texts with varied quality. In this paper, we propose a novel perspective that considers the role of LLMs in RAG as ``Information Refiner'', which means that regardless of correctness, completeness, or usefulness of retrieved texts, LLMs can consistently integrate knowledge within the retrieved texts and model parameters to generate the texts that are more concise, accurate, and complete than the retrieved texts. To this end, we propose an information refinement training method named \textbf{\textsc{InFO}-RAG} that optimizes LLMs for RAG in an unsupervised manner. \modelname is low-cost and general across various tasks. Extensive experiments on zero-shot prediction of 11 datasets in diverse tasks including Question Answering, Slot-Filling, Language Modeling, Dialogue, and Code Generation show that \modelname improves the performance of LLaMA2 by an average of 9.39\% relative points. \modelname also shows advantages in in-context learning and robustness of RAG.

\end{abstract}

\section{Introduction}

\begin{figure}[t]
\centering
\includegraphics[width=\linewidth]{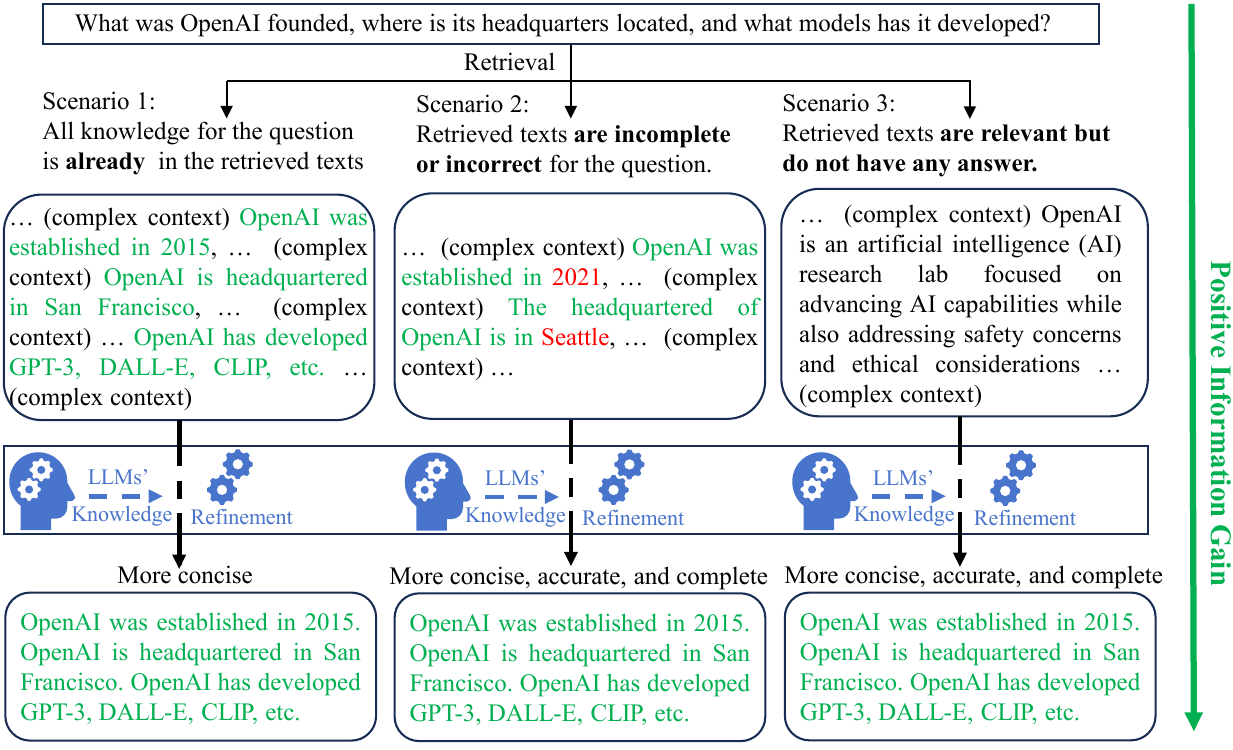} 
\caption{We consider the role of LLMs in RAG as ``Information Refiner'' that can generate more concise, accurate, and complete texts than the input retrieved texts. In this way, LLM can consistently make RAG system produce positive information gain.}
\label{motivation}
\end{figure}

Retrieval-augmented generation (RAG) is a popular framework in modern NLP systems that equips neural with retrieved information for text generation like open-domain question answering, dialogue~\cite{rag,realm} etc. Recently, RAG has been applied to large language models (LLMs) to provide additional knowledge and mitigate issues such as hallucination~\cite{peng2023check,shi2023replug,ren2023investigating}.

Despite the improved performance of retrieval models, the internet continues to be inundated with fake news, rumors, and fragmented, noisy information, posing challenges for retrieval models to reliably identify and shield against such content~\cite{sun2022rumor,beir}. Consequently, not all retrieved texts are beneficial, necessitating that LLMs determine how to judiciously utilize them. However, pre-training tasks do not explicitly enable LLMs to learn how to utilize the retrieved texts with varied quality for generation. For a question and its retrieved texts as input sequence, RAG aims to minimize the negative log-likelihood (NLL) of sub-sequence (question and generated answer) by referring to the retrieved texts. However, mainstream pre-training for LLMs with decoder-only architecture is language modeling based on the prefix~\cite{gpt,llama}, the training objective aims to minimize the negative log-likelihood (NLL) of the entire input sequence (retrieved texts, question, and generated answer)~\cite{mikolov2012statistical}. This gap causes LLMs to only regard the input retrieved texts as a part of the prefix for language modeling rather than additional reference, which leads to the following problems. Firstly, for the long and complex retrieved texts, LLMs struggle to extract the correct answers~\cite{deng2023regavae} accurately. Secondly, in situations where the retrieved texts cannot address the task, LLMs lack the capability to integrate the knowledge within model parameters with the retrieved texts to generate improved texts. Thirdly, LLMs are susceptible to incorrect and noisy information in retrieved texts, posing a risk of being misled~\cite{benchmark,yoran2023making}.

To solve above problems, some previous methods explore strategies for how or when to perform retrieval for LLMs by prompt techniques~\cite{self-ask,dsp,searchain,asai2023self}. However, prompt cannot materially change the ability of LLMs to utilize retrieved texts because model parameters are not updated for this ability. Some methods fine-tune LLMs on the constructed RAG data for a specific task such as QA~\cite{yoran2023making,yu2023chainofnote}. However, under the trend that LLMs are regarded as foundation models for various tasks in zero-shot setting, fine-tuning LLMs only on a few tasks make LLMs limited to the RAG of training tasks and lose their generalizability. Because catastrophic forgetting still exists in supervised fine-tuning of LLMs~\cite{cf}. Although constructing data for a large number of tasks can alleviate this, it is hard to design the data in various RAG tasks and requires high data annotation costs. Our paper aims to fundamentally improve the ability of LLMs to utilize retrieved texts while preserving the generalizability of LLMs for various RAG tasks in zero-shot setting, which is orthogonal to prompt techniques and can be combined with them to get better performance.

In this paper, considering that LLMs have a certain ability to use their own knowledge to examine information~\cite{cov}, we introduce a novel perspective to reassess the role of LLMs in RAG. Specifically, we propose considering LLMs as \textbf{``Information Refiner''}. The key idea behind this is to continue training the pre-trained LLMs with an Information Refinement objective that regardless of the correctness, completeness, or usefulness of the input retrieved texts, LLMs can consistently integrate knowledge within the retrieved texts and model parameters to generate the texts that are more concise, accurate, and complete than the retrieved texts (Figure~\ref{motivation}). We term this process ``Positive Information Gain''. This enables LLMs to extract correct information from complex texts as well as resist and rectify retrieved erroneous information and noise, thereby improving the information bottleneck of the RAG and allowing the knowledge capacity of RAG to approximate the combined knowledge of IR and LLMs.

We make the information refinement training work in a completely unsupervised manner, such that it is easy to obtain large-scale training data and maintain the generalizability of the trained LLMs that can be used in various RAG tasks in zero-shot setting. Specifically, we propose an unsupervised training method named \modelnamens. \modelname classifies the retrieved texts into three scenarios (shown in Figure~\ref{motivation}) and proposes the unsupervised training task for each scenario. For the first scenario that all knowledge for the question is already in the retrieved texts, LLMs need to accurately extract relevant knowledge from complex retrieved texts and generate more concise texts. For the second scenario that retrieved texts are incomplete or incorrect for the question, LLMs need to combine the knowledge within model parameters to verify the retrieved texts, correct the wrong knowledge, and complete the missing knowledge. For the third scenario that retrieved texts are relevant but do not have any answer, LLMs need to find the knowledge within model parameters based on relevant context to generate correct answers. We mix the above three tasks to train \modelname unsupervisedly.

Main contributions of this paper are as follows:

\noindent (1) We introduce a novel perspective to reassess the role of LLMs in the RAG system that considers LLMs as \textbf{``Information Refiner''} that can produce positive information gain in RAG scenarios.

\noindent (2) We propose an unsupervised training method named \modelname that enables LLMs to perform information refinement in RAG. \modelname is low-cost and general for various RAG tasks.

\noindent (3)  Extensive experiments show \modelname enhances the zero-shot RAG of LLaMA2 across Question Answering, Slot-Filling, Language Modeling, Dialog, and Code Generation. \modelname also shows advantages in in-context learning and robustness of RAG. Code is released at \url{https://github.com/xsc1234/INFO-RAG/}.

\section{Related Work}

\paragraph{Retrieval Augmented Generation} 

Retrieval augmented generation (RAG) aims to provide additional knowledge for language models by retrieving information from external databases~\cite{rag,realm,retro,atlas}. RAG makes the text generated by LLM more accurate and credible, and is widely used in Open-domain QA~\cite{dpr,trivedi2022interleaving}, dialogue~\cite{cai2018skeleton,cai2019retrieval} and Code Generation~\cite{codeg}. Recently, RAG has also been widely applied in LLMs~\cite{peng2023check,shi2023replug,ren2023investigating}. The form of RAG in LLMs is using the retrieved texts as contexts~\cite{ram2023context}. 

Some studies have noted that noise in retrieved texts will interfere with the performance of the language model or even mislead it~\cite{searchain,wang2023self,benchmark,xu2024list}. These works try to solve this problem from the interactive framework between IR and LM, while our work points out a more essential view. That is, previous studies on RAG do not define the role of LLMs in RAG clearly. Our paper introduces a novel perspective to reassess the role of LLMs in RAG that considers LLMs as ``Information Refiner''.

\paragraph{Unsupervised Learning of RAG}

Unsupervised learning of RAG can be divided into the training of retrievers and language models. As for retrievers, REALM~\cite{realm} proposes using masked language modeling to pre-train a knowledge retriever. REPLUG~\cite{shi2023replug} trains the retriever according to the feedback from black-box LM. As for language models, RETRO~\cite{retro} improves language models by retrieving tokens. Atlas proposes pretext tasks to jointly train the retriever and language model. However, these two methods focus on the model of encoder-decoder architecture, which is inconsistent with the current mainstream LLMs based on decoder-only.

Previous unsupervised training methods do not consider the specific role that language models should play in RAG. In this paper, we focus on training language model as an ``Information Refiner'' that can further improve the information bottleneck of RAG and be robust to retrieved texts.

\section{Our \modelname}

This section introduces our \textbf{\modelnamens}, an unsupervised training method to enable LLMs to perform information refinement in RAG. Firstly, we summarize the retrieved texts in RAG into three scenarios and define the positive information gain for each scenario. Secondly, we construct sample pairs in which the output has information gain compared to the input for these three scenarios and design three training tasks. Thirdly, we train LLMs under our designed tasks on the unsupervised samples. Unsupervised training makes \modelname low-cost and general for RAG in various tasks.

\subsection{Positive Information Gain in RAG} \label{scenario}

In this paper, we introduce a novel perspective to reassess the role of LLMs in RAG that LLMs should be the ``Information Refiner'' that can produce ``Positive Information Gain'' in the information flow of RAG. This section details the scenarios of retrieved texts and defines specific information gain LLMs should produce in each scenario.

\textbf{Scenario 1.} The first scenario is that all knowledge for the question is already in the retrieved texts. Even if the correct knowledge already exists in the retrieved texts, complex and lengthy retrieved texts are not conducive for users to directly obtain the knowledge. Therefore, the positive information gain in this scenario means that LLMs extract correct knowledge as much as possible while removing irrelevant information, thereby generating more direct and concise texts for users.

\textbf{Scenario 2.} The second scenario is that although the retrieved texts contain some usable knowledge, they still contain some incomplete or incorrect knowledge. This scenario is very common, especially with the current proliferation of fake news, misinformation, and fragmented knowledge on the Internet. There has been study proving that noise and erroneous knowledge in retrieved texts greatly mislead the generation of LLMs~\cite{searchain}. The positive information gain in this scenario is that LLMs can exploit the knowledge within their parameters to verify the knowledge in the retrieved texts. Utilize accurate knowledge, rectify incorrect knowledge, and complete missing knowledge

\textbf{Scenario 3.} The third scenario is that the retrieved texts do not have any answer that can used to solve the question. This scenario means that the question is very difficult or the target knowledge is very long-tail for information retrieval systems. Even in this case, the retrieval model's ability to model semantics allows it to provide texts that are semantically related to the question~\cite{dpr}. Therefore, the positive information gain in this scenario is that LLMs can stimulate the knowledge within their parameters based on semantically relevant context to solve the question.

\begin{figure*}[t]
\centering
\includegraphics[width=\linewidth]{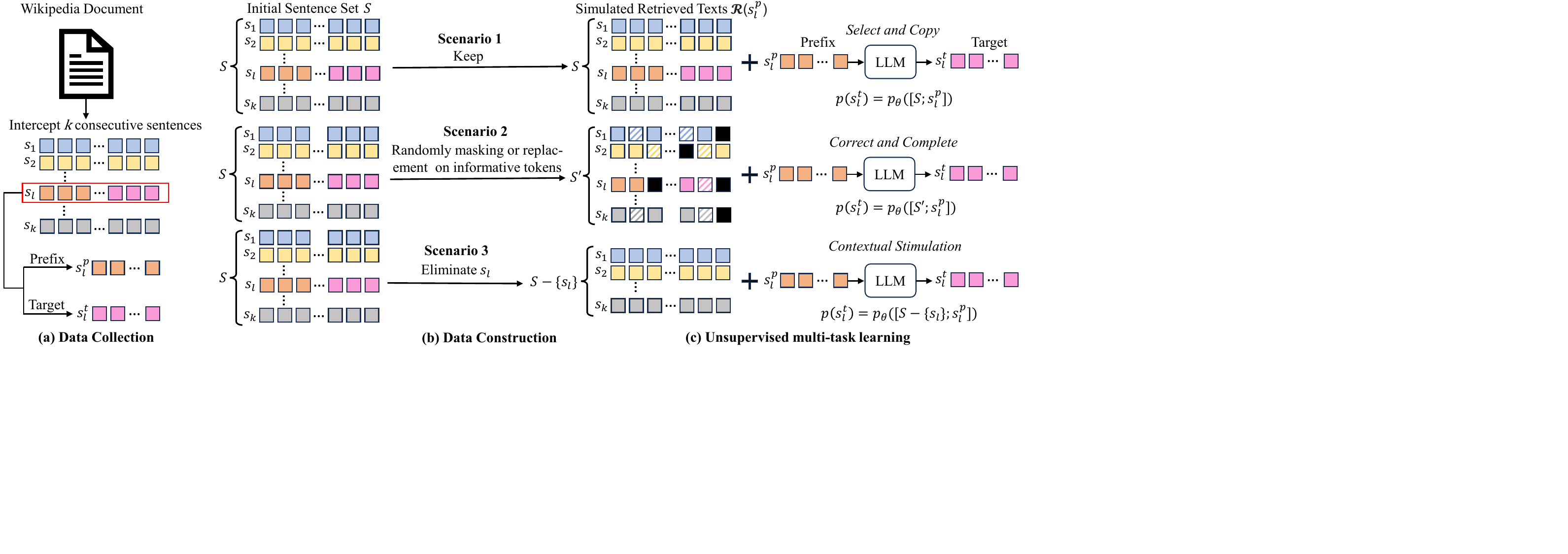} 
\caption{Overview of our \modelnamens. Each sample is only processed for a single scenario to avoid data leakage.}
\label{method}
\end{figure*}

\subsection{Unsupervised Learning}
This section introduces unsupervised learning in \modelnamens. We construct the input-output pairs that satisfy the information gain in the above three scenarios on Wikipedia. We continue to train pre-trained LLMs on the constructed data to perform information refinement in the form of next token prediction in prefix language modeling, which is general for various tasks. Pipeline is in Figure~\ref{method}.

\subsubsection{Data Collection}
The data construction is performed on English Wikipedia. Specifically, for each document $d$ in Wikipedia, we intercept $k$ consecutive sentences from $d$ and get the sentence set $S = [s_1,s_2,...,s_k]$. Our method randomly selects $s_l$ from $S$ and uses it as the object for language modeling. The first $\frac{1}{3}$ to $\frac{2}{3}$ of the tokens of $s_l$ are randomly intercepted as the prefix ($s^p_l$) and the other tokens of $s_l$ are used as the prediction target ($s^t_l$). We also perform the process (Section~\ref{cons}) on sentence set $S$ so that it can be used to simulate the retrieved texts $\mathcal{R}(s^p_l)$ for prefix $s^p_l$ in three scenarios for conditioning the generation of $s^t_l$. Then, we can get an unsupervised training sample for prefix language modeling that predicts $s^t_l$ given the prefix $s^p_l$ and the retrieved texts $\mathcal{R}(s^p_l)$. This can be formulated as:
\begin{equation}
p(s^t_l) = p_{\theta}([\mathcal{R}(s^p_l);s^p_l]),
\end{equation}
$\theta$ are parameters of LLMs, $[\mathcal{R}(s^p_l);s^p_l]$ is the concatenation of $\mathcal{R}(s^p_l)$ and $s^p_l$ by a special token.

\subsubsection{Data Construction and Training Tasks} \label{cons}
This section details our data construction and training tasks for three scenarios in Section~\ref{scenario}. 

For \textbf{Scenario 1} that needs LLMs to extract the correct knowledge from the complex texts, we propose the training task named \textit{Select and Copy}. Specifically, given the sentence set $S$ for a sample, \textit{Select and Copy} directly uses all sentences in $S$ as retrieved texts for conditioning LLMs to predict $s^t_l$ for the given prefix $s^p_l$. This can be formulated as:
\begin{equation}
p(s^t_l) = p_{\theta}([S;s^p_l]).
\end{equation}
In \textit{Select and Copy}, $s_l$ (both $s^p_l$ and $s^t_l$) has been contained in the retrieved texts $S$, this needs LLMs to select the texts matching the prefix $s^p_l$ from the complex retrieved texts $S$ and directly copy the target $s^t_l$ for generation. The information gain between $s^t_l$ and input retrieved texts $S$ is that $s^t_l$ is more concise to be used as the postfix for $s^p_l$.

For \textbf{Scenario 2} that needs LLMs to verify the knowledge in the retrieved texts, utilize accurate knowledge, rectify incorrect knowledge, and complete missing knowledge. We propose the training task named \textit{Correct and Complete}. Given a sentence set $S$, firstly, this task uses the stability of word distribution between layers to get informative tokens. The intention for this is that the more unstable the word distribution of the token among the topmost layers is, the more it indicates that the token is an informative token. We follow~\cite{dola} to achieve this. Specifically, for each sentence $s_i$ in $S$, our method obtains the next word distribution of the $a\mbox{-}th$ token $s^{[a]}_i$ given prefix $s^{<a}_i$ of $s_i$ in each layer of LLM as:
\begin{equation}
d_{j}(s^{[a]}_i|s^{<a}_i) = \textrm{softmax}(\textrm{\textbf{W}}\textrm{\textbf{H}}^{[a]}_j),
\end{equation}
in which $j$ indicates the $j\mbox{-}th$ layer of LLMs, $\textrm{\textbf{H}}^{[a]}_j \in \mathbb{R}^{h}$ is the hidden states for token $s^{[a]}_i$ in the $j\mbox{-}th$ layer, $\textrm{\textbf{W}} \in \mathbb{R}^{h \times v}$ is the vocabulary head that maps the hidden states $\textrm{\textbf{H}}^{[a]}_j$ to the word distribution with vocabulary size $v$. Then, for the LLM with $N$ layers, our method uses Jensen-Shannon Divergence (JSD) to measure the differences in word distribution between layers and gets the word distribution stability of token $s^{[a]}_i$ as:
\begin{align}
   O^{[a]}_i = \max_{j\in J}\textrm{JSD}(d_N(s^{[a]}_i|s^{<a}_i)||d_j(s^{[a]}_i|s^{<a}_i)), \notag
\end{align}
in which $J$ is the set of candidate layers ($0\mbox{-}th$ to $\frac{N}{2}\mbox{-}th$ layers), $d_N(s^{[a]}_i|s^{<a}_i)$ is the word distribution of the last layer. This design aims to find the layer with the largest word distribution difference between the last layer and use the JSD of the two as the word distribution stability of the token $s^{[a]}_i$~\cite{dola}. For each token of $s_i$, we obtain its word distribution stability in parallel and get the set of word distribution stability for $s_i$ as:
\begin{equation}
\mathbb{O}_i = \{O^{[0]}_i,O^{[1]}_i,...,O^{[n]}_i\}.
\end{equation}
We choose the tokens corresponding to the top 50\% of the elements in $\mathbb{O}_i$ as informative tokens within the sentence $s_i$. Subsequently, we apply a specific percentage ($30\%$) of random masking and replacement to these tokens. For the randomly selected token, we replace it with [MASK] with a $50\%$ probability to simulate the incomplete knowledge, and randomly replace it with another token with a $40\%$ probability to simulate the incorrect knowledge, while keeping it unchanged with a $10\%$ probability to simulate the correct knowledge. We do the above pipeline for each sentence in the set $S$ and get the processed set $S'$. RAG in \textit{Correct and Complete} can be formulated as:
\begin{equation}
p(s^t_l) = p_{\theta}([S';s^p_l]).
\end{equation}
In \textit{Correct and Complete}, the broken $s_l$ with noise is already in $S'$. The information gain in this task requires LLM to extract, correct, and complete the knowledge in $s_l$ from $S'$ to generate $s^t_l$.

For \textbf{Scenario 3} that needs LLMs to find answers from their knowledge based on relevant texts in context. We propose the training task named \textit{Contextual Stimulation}. \textit{Contextual Stimulation} eliminates $s_l$ (both  $s^p_l$ and $s^t_l$) from the set $S$ and uses the remaining sentences as retrieved tests for generation, which can be formulated as:
\begin{equation}
p(s^t_l) = p_{\theta}([S-\{s_l\};s^p_l]).
\end{equation}
In \textit{Contextual Stimulation}, each sentence in retrieved texts $S-\{s_l\}$ is semantically relevant to $s^p_l$ but cannot help LLMs to directly generate $s^t_l$. LLMs need to be stimulated by relevant information to generate $s^t_l$ based on their own knowledge.

\subsubsection{Training Strategy}

After the data construction for three training tasks, we mix them for multi-task training. Specifically, we use LoRA~\cite{lora} to train the pre-trained LLMs on the mixed dataset of three tasks. Three tasks are trained alternately in batches. Since \textit{Select and Copy} is relatively simple for LLMs, it only accounts for $20\%$ of the batches, while \textit{Correct and Complete} and \textit{Contextual Stimulation} each account for $40\%$ of the batches. Using LoRA not only reduces training costs but also makes our method plug-and-play. The trained LoRA parameters are loaded when LLMs need to perform RAG and unloaded when RAG is not needed.

\section{Experiments}

\begin{table*}[t]
\centering
\renewcommand\arraystretch{1.1}
\setlength\tabcolsep{2.8
pt}
\scalebox{0.85}{
\begin{tabular}{lcccccccccccc}
\toprule
\multicolumn{1}{c}{} & \multicolumn{2}{c}{Soft-Filling}   & \multicolumn{2}{c}{ODQA}                        & \multicolumn{2}{c}{Multi-Hop QA}             & \multicolumn{1}{c}{LFQA} & \multicolumn{1}{c}{Dialog} & \multicolumn{1}{c}{LM}       & \multicolumn{2}{c}{Code Gen}              & \multirow{3}{*}{Overall} \\
\multicolumn{1}{c}{}  & \multicolumn{2}{c}{Accuracy}                  & \multicolumn{2}{c}{Accuracy}        & \multicolumn{2}{c}{Accuracy}             & \multicolumn{1}{l}{ROUGE} & \multicolumn{1}{c}{F1} & \multicolumn{1}{c}{ROUGE}       & \multicolumn{2}{c}{CodeBLEU}              & \\ 
\multicolumn{1}{c}{}  & T-REx          & \multicolumn{1}{c}{ZS} & NQ             & \multicolumn{1}{l}{WebQ} & Hotpot         & \multicolumn{1}{l}{Musique} & \multicolumn{1}{c}{ElI5} & \multicolumn{1}{c}{Wow}    & \multicolumn{1}{c}{WikiText} & Python         & \multicolumn{1}{c}{Java} &                          \\ \hline
LLaMA-2-7B              & 55.60          & 54.08 & \textbf{46.82} & 43.52                                      & 39.40          & 25.95                        & 15.18                     & 7.85                        & 60.77                         & 21.44          & 22.99                     & 35.78                    \\
\quad + \modelname    & \textbf{65.91} & \textbf{57.01}   & 45.74          & \textbf{44.68}                   & \textbf{46.56} & \textbf{30.19}               & \textbf{17.18}            & \textbf{9.09}               & \textbf{62.91}                & \textbf{26.75} & \textbf{32.06}            & \textbf{39.83}           \\ \hline
LLaMA-2-7B-chat        & 60.63          & 55.03  & 49.42          & 46.72                                     & 50.03          & 42.69                        & 27.81                     & 10.21                       & 60.26                         & 22.46          & 23.90                     & 40.83                    \\
\quad + \modelname        & \textbf{65.77} & \textbf{58.32}  & \textbf{53.93} & \textbf{49.13}                    & \textbf{52.01} & \textbf{44.45}               & \textbf{28.15}            & \textbf{10.49}              & \textbf{63.24}                & \textbf{27.25} & \textbf{28.79}            & \textbf{43.78}           \\ \hline
LLaMA-2-13B      & 60.08          & 50.77  & 47.40          & 44.62       & 42.12          & 25.78                        & 14.80                     & 7.04                        & 62.20                         & 21.52          & 29.16                     & 36.86                    \\
\quad + \modelname   & \textbf{62.80} & \textbf{55.63}   & \textbf{47.82} & \textbf{45.42}     & \textbf{51.48} & \textbf{35.02}               & \textbf{17.48}            & \textbf{7.20}               & \textbf{64.14}                & \textbf{29.00} & \textbf{35.50}            & \textbf{41.04}           \\ \hline
LLaMA-2-13B-chat   & 62.53          & 56.81       & 50.36          & 45.47      & 61.23          & 47.06                        & 27.07                     & 11.19                       & 60.52                         & 22.34          & 30.96                     & 43.23                    \\
\quad + \modelname        & \textbf{65.39} & \textbf{59.05}    & \textbf{54.04} & \textbf{51.07}                  & \textbf{61.91} & \textbf{47.93}               & \textbf{27.24}            & \textbf{11.38}              & \textbf{63.92}                & \textbf{31.98} & \textbf{38.12}            & \textbf{46.55}           \\ \toprule
\end{tabular}
}
\caption{Overall performance on retrieval-augmented generation on 11 datasets across 7 tasks in zero-shot setting.}
\label{res_overall}
\end{table*}

\subsection{Datasets and Evaluation Metrics}
 To demonstrate the generality of our unsupervised training method, we evaluate the performance of \modelname on eleven datasets across seven tasks.

 \noindent \textbf{Open-domain Question Answering} Open-domain QA is a typical knowledge-intensive task that can directly evaluate the knowledge of LLMs. We use Natural Questions~\cite{nq} (NQ) and WebQuestions~\cite{webq} (WebQ) as the datasets. We use cover Exact Match (EM) to determine whether the ground truth exactly appears in the output and the accuracy is used as the evaluation metric, following~\cite{schick2023toolformer}

 \noindent \textbf{Soft Filling} Soft filling requires LLMs to output the object entities for the input subject entity and relation. We use two knowledge-intensive datasets including Zero Shot RE~\cite{zs} (ZS) and T-REx~\cite{trex}. We use the same evaluation metric as Open-domain QA.
  
 \noindent \textbf{Long-Form Question Answering} Compared with open-domain QA, LFQA is the QA task whose ground truth answer is a relatively long text. We use ELI5~\cite{eli5}, a knowledge-intensive dataset for LFQA. We use ROUGE-L as the evaluation metric~\cite{kilt}.

 \noindent \textbf{Dialogue} Dialogue in our experiment focuses on the factual knowledge. We use Wizard of Wikipedia~\cite{wow} (WoW), a knowledge-powered dialogue dataset whose conversation is grounded with knowledge. We use F1 as the evaluation metric~\cite{kilt}.
 
 \noindent \textbf{Language Modeling} We use WikiText-103~\cite{wikitext}, a popular dataset for language modeling. We use ROUGE-L as the evaluation metric.

 \noindent \textbf{Multi-Hop Question Answering} Multi-hop QA measures the ability of LLMs to perform combined reasoning on multiple knowledge. We use HotpotQA~\cite{hotpotqa} and Musique~\cite{musique} for this task. We use the same evaluation metric as Open-domain QA.

 \noindent \textbf{Code Generation} Code generation aims to generate the code for the given natural language. We use Java and Python in CodeXGLUE~\cite{codex} for this task. We use CodeBLEU~\cite{codebleu} as the evaluation metric. 

\subsection{Experimental Settings} 
LLMs in our paper include LLaMA-2-7B, 13B and their chat version~\cite{llama2}. We use LoRA to fine-tune these pre-trained LLMs on four A100 GPUs with the learning rate of 1e-5, per-gpu batch size of 4 (for 7B) and 2 (for 13B) for 5K steps. As for the training data, we intercept $15$ consecutive sentences for each example.

As for the retrieval model and retrieval database, for Open-domain QA, Soft Filling and Language Modeling, we use ColBERTv2~\cite{colbertv2}, a late-interaction model with excellent generalization ability as the retriever, and use Wikipedia consisting of 21,015,324 passages~\cite{dpr} as retrieval database. 
For Code Generation, we SCODE-R~\cite{codeg} as code retriever and use deduplicated source codes in CodeSearchNET~\cite{codesearch} as retrievel database. 
For all the above tasks, we give Top-5 retrieved passages to each example. For LFQA, Dialog, and Multi-Hop QA, we use the list of contextual passages provided in the datasets as the retrieved list (distractor setting). \textbf{In each experiment, all baselines and our method share the same retrieved documents.}

\subsection{Experimental Results}

\paragraph{Main Results (Zero-Shot Setting)}
Experimental results in Table~\ref{res_overall} show the improvement (the average is 9.39\%) of our method on the utilization of retrieved knowledge from four aspects.

\noindent \textbf{(1)} Short and Direct Knowledge. Our method can significantly improve the RAG performance of LLaMA on ODQA and Slot-Filling tasks. The answer in ODQA and Slot-Filling is short and direct, it can directly reflect the ability of LLMs to utilize the knowledge in retrieved texts.

\noindent \textbf{(2)} Reasoning on Multiple Knowledge. Our \modelname has advantages in cross-passage reasoning on multiple knowledge of retrieval lists. Questions in both HotpotQA and Musique are complex and need multiple knowledge from different passages. These questions not only require LLMs to extract correct knowledge from the retrieved passage list but also to combine the knowledge of different passages in the list for reasoning to give the final answer.

\noindent \textbf{(3)} Long and Complex Knowledge. Our \modelname can improve the RAG performance of LLaMA on LFQA, Dialogue and Language Modeling. These tasks require LLaMA to output long and complex texts grounded with intensive knowledge.

\noindent \textbf{(4)} Code Knowledge. 
Our \modelname can also improve the RAG performance of LLaMA on Code Generation. This further demonstrates the cross-task generality of \modelnamens. Our method is only trained on natural language but can also show advantages in programming language tasks, which demonstrates that \modelname successfully enables LLMs to learn how to exploit the retrieved information rather than just fitting the data. Unsupervised and prefix language modeling training paradigms make \modelname general in various tasks.

\paragraph{Results on In-context Learning for RAG}
Besides, our \modelname allows further improvement cooperating with in-context learning (ICL). ICL~\cite{iclorigin} works by prepending a few examples of the target task before the query, which helps LLMs understand the task.
However, ICL may not always help in the RAG setting, mainly due to the confusion between the retrieved texts of the query and the few-shot examples.
As shown in Table~\ref{icl_res}, LLaMA-2 cannot further improve the RAG performance from ICL, even sometimes hurt by the few-shot examples while \modelname can further improve RAG by ICL. This is mainly because \modelname enables LLaMA to understand the task form of RAG, thereby better learning the general task pattern from ICL examples. In this experiment, we construct the ICL example consisting of a query, a relevant passage, and an answer. For a fair comparison, we need to ensure that the performance of our method and the baseline are close in non-ICL setting. Therefore, we select queries for which the baseline gives the same answer as our method (both correct or both incorrect) and evaluate the ICL performance on these queries. 

\begin{table*}[t]
\centering
\renewcommand\arraystretch{1.05}
\setlength\tabcolsep{2
pt}
\scalebox{0.85}{
\begin{tabular}{lllllllllllll}
\toprule
{}      & \multicolumn{3}{c}{T-REx}    & \multicolumn{3}{c}{ZS}  & \multicolumn{3}{c}{NQ}                                              & \multicolumn{3}{c}{WebQ}                              \\ 
{}                 & has-ans.       & replace        & \multicolumn{1}{l}{no-ans.}       & has-ans.       & replace        & \multicolumn{1}{l}{no-ans.}        & has-ans.       & replace        & \multicolumn{1}{l}{no-ans.}       & has-ans.       & replace        & no-ans.       \\ \hline
{LLaMA-2-7B}       & 67.19          & 38.37          & {6.49}          & 64.41          & 12.78          & 2.44    & \textbf{65.54} & 16.91          & {3.41}          & 60.64          & 25.68          & {7.90}                \\
{\quad + \modelname }      & \textbf{79.80} & \textbf{41.79} & {\textbf{7.04}} & \textbf{68.10} & \textbf{13.55} & \textbf{3.26}   & 64.43          & \textbf{22.68} & {\textbf{4.70}} & \textbf{62.70} & \textbf{26.48} & {\textbf{8.96}} \\ \hline
{LLaMA-2-7B-chat}  & 73.79          & 40.56          & {4.87}          & 66.71          & 14.19          & 1.63   & 68.72          & 20.81          & {4.50}          & 66.86          & 28.63          & {5.62}                 \\
{\quad + \modelname }    & \textbf{80.01} & \textbf{42.92} & {\textbf{5.42}} & \textbf{69.64} & \textbf{15.02} & \textbf{2.65} & \textbf{70.99} & \textbf{23.14} & {\textbf{5.62}} & \textbf{68.73} & \textbf{29.74} & {\textbf{9.12}} \\ \hline
{LLaMA-2-13B}     & 72.26          & 39.47          & {7.76}          & 60.14          & 19.71          & 4.69   & \textbf{65.94} & 18.45          & {4.42}          & 62.09          & 26.63          & {9.27}           \\
{\quad + \modelname }       & \textbf{75.80} & \textbf{44.08} & {\textbf{8.48}} & \textbf{65.94} & \textbf{23.21} & \textbf{4.90} & 64.98          & \textbf{27.60} & {\textbf{8.02}} & \textbf{63.51} & \textbf{28.24} & {\textbf{9.88}} \\ \hline
{LLaMA-2-13B-chat}  & 75.96          & 43.79          & {5.59}          & 67.03          & 16.58          & 1.42  & 69.37          & 30.72          & {6.16}          & 65.07          & 31.88          & {5.47}                 \\
{\quad + \modelname }     & \textbf{79.25} & \textbf{48.59} & {\textbf{6.67}} & \textbf{70.26} & \textbf{25.02} & \textbf{3.87} & \textbf{73.73} & \textbf{33.85} & {\textbf{8.39}} & \textbf{70.59} & \textbf{37.48} & {\textbf{11.25}} \\ \toprule

\end{tabular}
}
\caption{Experimental results on three scenarios. ``has-ans.'' is the first scenario that correct answers are in retrieved texts. ``replace'' is the second scenario that correct answers are randomly replaced with other phrases to simulate the incorrect and incomplete knowledge. ``no-ans.'' is the third scenario that retrieval cannot find any answers.}
\label{res_scenario}
\end{table*}

\begin{table}[t]
\centering
\renewcommand\arraystretch{1.1}
\setlength\tabcolsep{2.5
pt}
\scalebox{0.80}{
\begin{tabular}{llcccccc}
\toprule
\multirow{2}{*}{Data} & \multirow{2}{*}{Model} & \multicolumn{6}{c}{Number of Examples in ICL}                                              \\
                          &                         & 0     & 2              & 4              & 8              & 12             & 16             \\ \hline
\multirow{2}{*}{NQ}       & LLaMA-2               & 43.36 & 23.34          & 16.60          & 39.22          & 44.32          & 43.00          \\
                          & +\modelname                & 43.36 & \textbf{44.35} & \textbf{45.88} & \textbf{44.45} & \textbf{47.75} & \textbf{46.25} \\ \hline
\multirow{2}{*}{WebQ}     & LLaMA-2                & 43.20 & 18.36          & 9.40           & 36.71          & 44.80          & 44.81          \\
                          & +\modelname                & 43.20 & \textbf{48.03} & \textbf{49.82} & \textbf{48.25} & \textbf{47.86} & \textbf{47.29} \\ \hline
\multirow{2}{*}{T-REx}    & LLaMA-2                & 59.83 & 47.05          & 49.11          & 56.51          & 55.23          & 56.31          \\
                          & +\modelname                & 59.83 & \textbf{63.08} & \textbf{63.45} & \textbf{63.54} & \textbf{63.57} & \textbf{63.38} \\ \hline
\multirow{2}{*}{ZS}       & LLaMA-2                & 52.41 & 42.71          & 37.05          & 50.40          & 50.20          & 51.01          \\
                          & +\modelname                & 52.41 & \textbf{56.53} & \textbf{60.37} & \textbf{59.86} & \textbf{59.75} & \textbf{59.85} \\ \toprule
\end{tabular}
}
\caption{RAG performance changes with number of examples in In-context learning.}
\label{icl_res}
\end{table}

\begin{table}[t]
\centering
\renewcommand\arraystretch{1.05}
\setlength\tabcolsep{6.5
pt}
\scalebox{0.80}{
\begin{tabular}{lllll}
\toprule
              & \multicolumn{2}{c}{Multi-Hop QA} & \multicolumn{2}{l}{Slot-Filling} \\
              & HotpotQA        & Musique      & T-REx     & zsRE         \\ \hline
Previous SOTA & 28.19           & 10.03           & 63.10           & 57.09          \\
SearChain     & 31.21           & 11.27          &  64.58          & 58.91          \\
+ \modelname  & \textbf{33.04}           & \textbf{12.10}          & \textbf{66.95}           & \textbf{60.72}   \\ \toprule     
\end{tabular}
}
\caption{Enhancement to the state-of-the-art RAG framework. Previous SOTA includes DSP, Self-Ask, React.}
\label{adaptive_sota_rag}
\end{table}

\paragraph{Enhancing Previous SOTA in Open-Retrieval Setting} 
We further show that our \modelname can cooperate well with the recent prompting techniques that perform multi-step reasoning to combine with retrieval to solve questions~\cite{searchain,dsp,self-ask,react}.
To make a fair comparison, we follow SearChain~\cite{searchain} that runs on Multi-Hop QA and Slot-Filling in open-retrieval setting that retrieves passages from the full Wikipedia in each reasoning step. SearChain and other baselines use LLaMA-2-13B-chat as the backbone. Then, we perform SearChain based on LLaMA-2-13B-chat trained by \modelname to show the enhancement to SearChain by \modelnamens. Results in Table~\ref{adaptive_sota_rag} show that \modelname can make SearChain achieve better performance. This provides additional support that our unsupervised \frameworkname training fundamentally improves the RAG performance of LLMs.

\subsection{Analysis}
\paragraph{Fine-grained Analysis for Three Scenarios}
As shown in Table~\ref{res_scenario}, our \modelname is effective in all three RAG scenarios and shows better robustness to incorrect, incomplete, and noisy retrieved texts. We propose corresponding unsupervised training tasks for the three scenarios of RAG. This section introduces the fine-grained analysis for each scenario. For \textbf{Scenario 1}, we use cover EM to select those samples that already contain the correct answers in the retrieval list. For \textbf{Scenario 2}, we randomly replace the correct answers in the retrieved texts with another phrase with the same properties. For \textbf{Scenario 3}, we use cover EM to select those samples that retrieved texts do not contain any correct answers. We count the accuracy of LLaMA on samples of these three scenarios respectively. Questions in the third scenario are more difficult than in the second scenario because retrieval models cannot find anything to solve them. Table~\ref{res_scenario} indicates that our method shows advantages in each scenario and is more robust regardless of whether the retrieved texts contain the correct answer.

\begin{table*}[t]
\centering
\renewcommand\arraystretch{1.0}
\setlength\tabcolsep{4.2
pt}
\scalebox{0.82}{
\begin{tabular}{lcccccccccccc}
\toprule
\multicolumn{1}{c}{} & T-REx          & \multicolumn{1}{c}{ZS}  & NQ             & \multicolumn{1}{l}{WebQ}  & Hotpot         & \multicolumn{1}{l}{Musique} & \multicolumn{1}{c}{ElI5} & \multicolumn{1}{c}{Wow}    & \multicolumn{1}{c}{WikiText} & Python         & \multicolumn{1}{c}{Java} &  Overall                        \\ \hline
LLaMA-2 w/o RAG  & 35.60          & 10.99      & 32.67          & 39.13                                & 29.16          & 5.83                        & 26.05                   & 10.71                       & 41.80                         & 20.67         & 25.87                     & 25.32                    \\
LLaMA-2 w/ RAG   & 62.53          & 56.81    & 50.36          & 45.47         & 61.23          & 47.06                        & 27.07                     & 11.19                       & 60.52                         & 22.34          & 30.96                     & 43.23                    \\
\quad + training on wiki & 62.55          & 56.79  & 49.23          & 45.05                            & 61.00          & 46.95                        & 26.31                    & 11.05                       & 60.84                         & 22.05          & 30.28                     & 42.92                    \\
\quad + \modelname     & \textbf{65.39} & \textbf{59.05}   & \textbf{54.04} & \textbf{51.07}                  & \textbf{61.91} & \textbf{47.93}               & \textbf{27.24}            & \textbf{11.38}              & \textbf{63.92}                & \textbf{31.98} & \textbf{38.12}            & \textbf{46.55}           \\ \toprule
\end{tabular}
}
\caption{Analysis on the best-performed model LLaMA-2-13B-chat.}
\label{training_on_wiki}
\end{table*}

\begin{table}[t]
\centering
\renewcommand\arraystretch{1.0}
\setlength\tabcolsep{1.5
pt}
\scalebox{0.82}{
\begin{tabular}{lllll}
\toprule
\multirow{2}{*}{Method}                             & \multicolumn{4}{c}{NQ}                  \\
                                    & original & has-ans. & replace & no-ans. \\ \hline
Baseline                            & 50.36         & 69.37         & 30.72        & 6.16        \\ 
S1: \textit{Select and Copy}        & 48.77         & 69.59         & 25.40     & 0.11    \\
S2: \textit{Correct and Complete}   & 51.59         & 70.42       & 32.71      & 4.48                \\
S3: \textit{Contextual Stimulation} & 52.75         & 72.50       & 31.77      & 8.86              \\
S2\&S3                            & 53.73         &  73.01       &  32.50     & 9.01              \\
\modelname (S1\& S2\&S3)             & 54.04        & 73.73        & 33.85       & 8.39             \\  \toprule
\end{tabular}
}
\caption{Effects of three training tasks.}
\label{effect_of_training_task}
\end{table}

\paragraph{Ablation Study} We conduct ablation study to explore the effects of the following factors.

\noindent \textbf{(1) Additional Training on Wikipedia. }We study whether our improvement is from helping the model to achieve information refinement, or simply because of additional training on Wikipedia. To this end, we train LLaMA-2 on Wikipedia with standard language modeling objective, by setting the same hyperparameters as our \modelnamens. The results in Table~\ref{training_on_wiki} show that this baseline leads to no improvement over the backbone LLaMA-2, confirming the effectiveness of our training method rather than additional training on Wikipedia.

\noindent \textbf{(2) Training tasks. }We perform three training tasks proposed in \modelname separately on original data and data constructed for each scenario to explore their effects respectively. Table~\ref{effect_of_training_task} shows that both S2 and S3 have gains in their scenarios. Although S1 has negative effects when performed alone, it can achieve the best results when trained together with S2 and S3. This is mainly because S1 alone is so simple that causes LLM to overfit the data. Adding S2 and S3 allows LLM to learn the task paradigm of information refinement, making LLM better extract the correct answer for \textbf{Scenario 1}.

\paragraph{Robustness to Retrieval Results}
Table~\ref{robutst} shows \modelname is more robust to changes in retrieval results including the ratio and position of positive passages and number of retrieved passages. More details can be found in Section~\ref{sec:appendix} of Appendix.

\noindent \textbf{Avoid Catastrophic Forgetting }Experiment on MMLU~\cite{mmlu} without RAG shows that \modelname performs very close to the original LLaMA-2 (7B: 45.0 vs. 45.3; 13B: 54.3 vs. 54.8), which indicates that \modelname enhances RAG while avoiding catastrophic forgetting. More details can be found in Section~\ref{exp_mmlu} of Appendix.

\begin{table}[t]
\centering
\renewcommand\arraystretch{1.05}
\setlength\tabcolsep{5.3
pt}
\scalebox{0.82}{
\begin{tabular}{lllll}
\toprule
\multirow{2}{*}{Datasets} & \multirow{2}{*}{Method} & \multicolumn{1}{l}{Max $\Delta$} & Max $\Delta$         & Max $\Delta$         \\
                          &                         & ratio                         & position          & number            \\ \hline
\multirow{2}{*}{NQ}       & LLaMA-2                 & -51.94\%                      & -16.18\%          & -25.43\%          \\
                          & + \modelname                & \textbf{-43.48\%}             & \textbf{-15.80\%}  & \textbf{-17.25\%} \\ \hline
\multirow{2}{*}{WebQ}     & LLaMA-2                 & -50.57\%                      & \textbf{-5.63\%}  & -22.13\%          \\
                          & + \modelname                & \textbf{-45.48\%}             & -8.72\%           & \textbf{-11.91\%} \\ \hline
\multirow{2}{*}{T-REx}    & LLaMA-2                 & -46.57\%                      & -9.45\%           & -5.95\%           \\
                          & + \modelname                & \textbf{-44.38\%}             & \textbf{-8.61\%}  & \textbf{-2.99\%}  \\ \hline
\multirow{2}{*}{ZS}       & LLaMA-2                 & -59.25\%                      & -13.40\%          & -12.37\%          \\
                          & + \modelname                & \textbf{-50.08\%}             & \textbf{-11.11\%} & \textbf{-11.43\%} \\ \toprule
\end{tabular}
}
\caption{Maximum relative performance change caused by changes in retrieval results.}
\label{robutst}
\end{table}

\section{Conclusion}

This paper proposes a novel perspective to reassess the role of LLMs in RAG that considers LLMs as ``Information Refiner''. This means that regardless of the correctness, completeness, or usefulness of the retrieved texts, LLMs can consistently integrate knowledge within model parameters and the retrieved texts to generate texts that are more concise, accurate, and complete. To achieve it, we propose an information refinement training method named \modelname in an unsupervised manner, which is low-cost and general across various tasks. Extensive experiments across 11 datasets of 7 tasks in zero-shot setting show that \modelname improves the performance of LLMs for RAG. \modelname also shows advantages in ICL and robustness of RAG and can be combined with the SOTA RAG framework to further improve its performance.

\section*{Limitations}
This paper aims to enable LLMs to perform information refinement in RAG by unsupervised training, so as to accurately extract correct information and avoid the interference of incorrect information. The main limitation of this paper is that due to the lack of computing resources, we only conduct experiments on models with 7B and 13B parameter sizes. In the future, we consider using more computing resources to explore the performance of models with larger parameter sizes.

\section*{Ethics Statement}
After careful consideration, we believe that our paper does not introduce additional ethical concerns. We declare that our work complies with the \href{https://www.aclweb.org/portal/content/acl-code-ethics}{ACL Ethics Policy}.

\section*{Acknowledgements}
This work was supported by the National Key R\&D Program of China (2022YFB3103700, 2022YFB3103704), the National
Natural Science Foundation of China (NSFC) under Grants No. 62276248 and U21B2046, and the Youth Innovation Promotion Association
CAS under Grants No. 2023111.

\bibliography{anthology,custom}

\appendix

\section{More Analysis}
\label{sec:appendix}

\subsection{Robustness to ratio of Positive Passages}

\begin{table*}[h]
\centering
\renewcommand\arraystretch{1.1}
\setlength\tabcolsep{12
pt}
\scalebox{0.9}{
\begin{tabular}{lllllllll}
\toprule
\multirow{2}{*}{Data} & \multirow{2}{*}{Model}  & \multicolumn{6}{c}{ratio of Positive Passages}      & \multirow{2}{*}{Max $\Delta$}                                              \\
                       &          & 100\%          & 80\%           & 60\%           & 40\%           & 20\%           & 0\%            \\ \hline
\multirow{2}{*}{NQ}    & LLaMA-2 & 88.11          & 82.71          & 80.81          & 77.62          & 69.73          & 42.35  & -51.94\%        \\
                       & + \modelname & \textbf{90.31} & \textbf{83.72} & \textbf{81.72} & \textbf{79.72} & \textbf{71.52} & \textbf{51.04} & \textbf{-43.48\%} \\ \hline
\multirow{2}{*}{WebQ}  & LLaMA-2 & 79.41          & 75.43          & 71.63          & 65.53          & 63.39          & 39.25 & -50.57\%         \\
                       & + \modelname & \textbf{83.66} & \textbf{76.23} & \textbf{74.23} & \textbf{69.05} & \textbf{65.74} & \textbf{45.61} & \textbf{-45.48\%} \\ \hline
\multirow{2}{*}{T-REx} & LLaMA-2 & 80.01          & 70.05          & 71.52          & 68.53          & 66.23          & 42.75    & -46.57\%      \\
                       & + \modelname & \textbf{83.52} & \textbf{73.22} & \textbf{74.93} & \textbf{72.32} & \textbf{70.12} & \textbf{46.45} & \textbf{-44.38\%} \\ \hline
\multirow{2}{*}{ZS}    & LLaMA-2 & 69.52         & 65.48         & 63.81          & 60.95          & 57.14          & 28.33   & -59.25\%       \\
                       & + \modelname & \textbf{72.50} & \textbf{72.62} & \textbf{67.62} & \textbf{67.86} & \textbf{60.48} & \textbf{36.19} & \textbf{-50.08\%} \\ \toprule
\end{tabular}
}
\caption{RAG performance changes with the ratio of positive passages (randomly select 500 samples).}
\label{ir_per}
\end{table*}

Our \modelname improves the robustness of RAG performance to retrieval performance. The performance of the retriever greatly affects the performance of LLM in RAG~\cite{benchmark}. We explore this in this section. Specifically, we simulate changes in retrieval performance by varying the ratio of positive and negative passages in the retrieved list and report the RAG performance with different ratios. Table~\ref{ir_per} shows \modelname performs better when the ratio is low and the performance is more stable than baseline when the ratio changes from 100\% to 0\% (Max $\Delta$). The model in this experiment is LLaMA-2-13B-chat.

\subsection{Robustness to Positive Passage Position}
Experimental results in Table~\ref{position} show that our \modelname consistently outperforms the baseline (LLaMA-2) regardless of where the positive passage (passage contains the correct answers) appears in the retrieved list. Specifically, we mix positive and negative passages in a ratio of 1:9 to simulate the retrieved passage list, vary the position of the positive passage in the retrieved list from 0 to 9, and evaluate the corresponding RAG performance respectively. The model in this experiment is LLaMA-2-13B-chat. Experimental results show that our \modelname not only outperforms the baseline at every position but also achieves more stable performance varying with the position (Max $\Delta$).

\begin{table*}[t]
\centering
\renewcommand\arraystretch{1.1}
\setlength\tabcolsep{4.5
pt}
\scalebox{0.9}{
\begin{tabular}{lllllllllllll}
\toprule
\multirow{2}{*}{Datasets} & \multirow{2}{*}{Method} & \multicolumn{10}{c}{Position of Positive Passage} & \multirow{2}{*}{Max $\Delta$}                                                                          \\
                          &                         & 0     & 1     & 2              & 3     & 4              & 5     & 6              & 7     & 8              & 9              \\ \hline
\multirow{2}{*}{NQ}       & LLaMA-2                & 54.94 & 48.05 & 46.05          & 46.45 & 46.35          & 48.30 & 48.35          & 47.15 & 51.64          & 50.44   & -16.18\%       \\
                          & + \modelname                & \textbf{63.23} & \textbf{58.34} & \textbf{54.54} & \textbf{54.44} & \textbf{53.54} & \textbf{53.24} & \textbf{53.84} & \textbf{54.44} & \textbf{53.34} & \textbf{53.34} & \textbf{-15.80\%} \\ \hline
\multirow{2}{*}{WebQ}     & LLaMA-2                 & 66.13 & 63.21 & 62.54          & 62.68 & 64.01          & 62.41 & 63.21          & 64.54 & 63.87          & 64.14  & \textbf{-5.63\%}       \\
                          & + \modelname                & \textbf{71.58} & \textbf{68.39} & \textbf{66.26} & \textbf{65.34} & \textbf{67.19} & \textbf{65.73} & \textbf{65.73} & \textbf{65.81} & \textbf{65.54} & \textbf{66.72} & -8.72\%\\ \hline
\multirow{2}{*}{T-REx}    & LLaMA-2                 & 64.43 & 60.13 & 58.34          & 60.23 & 58.54          & 59.14 & 59.74          & 60.53 & 63.53          & 63.23  & -9.45\%       \\
                          & + \modelname                & \textbf{70.72} & \textbf{66.23} & \textbf{64.93} & \textbf{65.23} & \textbf{65.43} & \textbf{64.83} & \textbf{66.03} & \textbf{67.23} & \textbf{64.63} & \textbf{66.83} & \textbf{-8.61\%} \\ \hline
\multirow{2}{*}{ZS}       & LLaMA-2                 & 63.04 & 59.04 & 54.59          & 55.03 & 55.17          & 57.15 & 56.42          & 57.89 & 58.04          & 59.47    & -13.40\%      \\
                          & + \modelname                & \textbf{66.42} & \textbf{63.33} & \textbf{59.04} & \textbf{60.23} & \textbf{61.42} & \textbf{61.66} & \textbf{60.00} & \textbf{61.19} & \textbf{60.23} & \textbf{62.14} & \textbf{-11.11\%} \\ \toprule
\end{tabular}
}
\caption{RAG performance changes with the position of positive passage (randomly select 500 samples).}
\label{position}
\end{table*}

\begin{table*}[t]
\centering
\renewcommand\arraystretch{1.1}
\setlength\tabcolsep{4.5
pt}
\scalebox{0.9}{
\begin{tabular}{lllllllllllll}
\toprule
\multirow{2}{*}{Datasets} & \multirow{2}{*}{Method} & \multicolumn{10}{c}{Number of Retrieved Passages}  & \multirow{2}{*}{Max $\Delta$}                                                                      \\
                          &                         & 1     & 2              & 3     & 4              & 5     & 6              & 7     & 8              & 9     & 10              \\ \hline
\multirow{2}{*}{NQ}       & LLaMA-2                 & 38.80 & 43.21 & 46.62         & 47.84 & 48.61          & 49.42 & 52.03          & 50.23 & 50.40          & 50.20    & -25.43\%      \\
                          & + \modelname                & \textbf{45.18} & \textbf{46.80} & \textbf{51.44} & \textbf{51.23} & \textbf{51.00} & \textbf{53.21} & \textbf{54.03} & \textbf{53.44} & \textbf{53.82} & \textbf{54.60} & \textbf{-17.25\%} \\ \hline
\multirow{2}{*}{WebQ}     & LLaMA-2                & 40.22 & 43.63 & 48.20     & 46.61 & 48.32          & 49.11 & 49.40          & 50.22 & 51.65          & 50.43   & -22.13\%       \\
                          & + \modelname                & \textbf{50.21} & \textbf{53.84} & \textbf{54.41} & \textbf{55.07} & \textbf{55.25} & \textbf{55.27} & \textbf{57.00} & \textbf{55.45} & \textbf{56.62} & \textbf{56.03} & \textbf{-11.91\%} \\ \hline
\multirow{2}{*}{T-REx}    & LLaMA-2               & 66.20 & 63.45 & \textbf{67.22}   & 64.45 & 64.43   & 65.40 & 64.41 & 65.22 & 63.22 & 65.01  & -5.95\%        \\
                          & + \modelname                & \textbf{66.25} & \textbf{66.03} & 66.31 & \textbf{65.80} & \textbf{67.23} & \textbf{67.22} & \textbf{66.65} & \textbf{67.83} & \textbf{67.03} & \textbf{67.40} & \textbf{-2.99\%} \\ \hline
\multirow{2}{*}{ZS}       & LLaMA-2                & 49.25 & 50.01 & 52.38    & 54.09 & 56.12   & 56.20 & 56.13          & 56.05 & 55.95    & 56.11   & -12.37\%       \\
                          & + \modelname                & \textbf{53.17} & \textbf{54.08} & \textbf{56.35} & \textbf{58.01} & \textbf{59.45} & \textbf{59.12} & \textbf{59.40} & \textbf{58.55} & \textbf{60.03} & \textbf{59.08} &\textbf{-11.43\%} \\ \toprule
\end{tabular}
}
\caption{RAG performance changes with the number of retrieved passages (randomly select 500 samples).}
\label{topk}
\end{table*}

\subsection{Robustness to Number of Retrieved Passages}
Experimental results in Table~\ref{topk} show that our \modelname consistently outperforms the baseline with the different number of retrieved passages (from 1 to 10) and is robust to the change of the number. In this experiment, we use LLaMA-2-13B-chat as the base model, change the number of retrieved passages from 1 to 10, and evaluate the corresponding performance.

\begin{table}[H]
\centering
\renewcommand\arraystretch{1.0}
\setlength\tabcolsep{4.9
pt}
\scalebox{0.95}{
\begin{tabular}{lllll}
\toprule
           &  T-REx          & ZS                                                            & NQ                                                            & WebQ                                                          \\ \hline
Baseline   & 51.47                                                         & 40.26          & 45.05          & 41.78          \\
+ INFO-RAG & \textbf{55.67} & \textbf{43.29} & \textbf{49.76} & \textbf{44.02} \\ \toprule
\end{tabular}
}
\caption{Works based on BM25.}
\label{bm25_as}
\end{table}

\subsection{Ablation Study on Masking Strategy}
\begin{table*}[t]
\centering
\renewcommand\arraystretch{1.05}
\setlength\tabcolsep{2
pt}
\scalebox{0.85}{
\begin{tabular}{lllllllllllll}
\toprule
{}      & \multicolumn{3}{c}{T-REx}    & \multicolumn{3}{c}{ZS}  & \multicolumn{3}{c}{NQ}                                              & \multicolumn{3}{c}{WebQ}                              \\ 
{}                 & has-ans.       & replace        & \multicolumn{1}{l}{no-ans.}       & has-ans.       & replace        & \multicolumn{1}{l}{no-ans.}        & has-ans.       & replace        & \multicolumn{1}{l}{no-ans.}       & has-ans.       & replace        & no-ans.       \\ \hline
{Baseline}     & 75.96	&43.79	&5.59	&67.03	&16.58	&1.42	&69.37	&30.72	&6.16	&65.07	&31.88	&5.47  \\
{Simple Mask}     & 78.43	&44.05	&5.75	&\textbf{70.30}	&19.45	&1.96	&73.59	&31.05	&6.51	&70.55	&32.96	&6.83 \\
{Our method}      &\textbf{79.25}	&\textbf{48.59}	&\textbf{6.67}	& 70.26	&\textbf{25.02}	&\textbf{3.87}	&\textbf{73.73}	&\textbf{33.85}	&\textbf{8.39}	&\textbf{70.59}	&\textbf{37.48}	&\textbf{11.25} \\ \toprule

\end{tabular}
}
\caption{Ablation study of masking strategy on three scenarios. ``has-ans.'' is the first scenario that correct answers are in retrieved texts. ``replace'' is the second scenario that correct answers are randomly replaced with other phrases to simulate the incorrect and incomplete knowledge. ``no-ans.'' is the third scenario that retrieval cannot find any answers.}
\label{abs_more}
\end{table*}
In general, Table~\ref{abs_mask} and~\ref{abs_more} show our masking strategy in Scenario 3 is more effective than simple and straightforward masking. Specifically, our method is more significantly effective in the scenarios that correct answers are randomly replaced with other phrases (replace) and retrieval cannot find any answers (no answer).

\subsection{Works with Different Retriever}
We evaluate our method and baseline (LLaMA2-13B-chat) with BM25 as the retriever, the experimental results shown in Table~\ref{bm25_as} indicate that our method still performs better than baseline when the retriever as BM25.

\begin{table}[H]
\centering
\renewcommand\arraystretch{1.0}
\setlength\tabcolsep{4.9
pt}
\scalebox{0.95}{
\begin{tabular}{lllll}
\toprule
{Method} & {T-REx} & {ZS}    & {NQ}    & {WebQ}  \\ \hline
{Baseline}        & {62.53}          & {56.81}          & {50.36}          & {45.47}          \\
{Simple Mask}     & {\textbf{64.05}} & {\textbf{58.91}} & {\textbf{53.80}} & {\textbf{50.55}} \\ \hline
{Our method}      & {\textbf{65.39}} & {\textbf{59.05}} & {\textbf{54.04}} & {\textbf{51.07}} \\ \toprule
\end{tabular}
}
\caption{Ablation study of masking strategy.}
\label{abs_mask}
\end{table}

\subsection{Performance on MMLU} \label{exp_mmlu}
Experimental results on MMLU benchmark in the setting without RAG shown in Table~\ref{mmlu} show that our \modelname significantly improves the performance of LLMs in RAG, while still maintaining its versatility and avoiding catastrophic forgetting. MMLU is a benchmark that measures massive multitask language understanding ability of LLMs. It covers 57 subjects across STEM, the humanities, the social sciences, and more. It ranges in difficulty from an elementary level to an advanced professional level, and it tests both world knowledge and problem-solving ability~\cite{mmlu}. Experiments show that our \modelname performs very close to the original LLaMA-2 on MMLU, which shows that our \modelname does not damage the basic language understanding ability of LLMs. This is mainly because the prefix language modeling training paradigm of our method is consistent with the pre-training task of LLMs. The difference is that in the training of prefix language modeling, our method learns to perform information refinement that utilizes the retrieved texts for the next token prediction.
\begin{table}[H]
\centering
\renewcommand\arraystretch{1.1}
\setlength\tabcolsep{2
pt}
\scalebox{0.65}{
\begin{tabular}{lccccc}
\toprule
                            & Humanities & STEM & Social-Sciences & Other & Average \\ \hline
LLaMA-2-7B w/o RAG                 & 42.9       & 36.4 & 51.2            & 52.2  & 45.3    \\
+ \modelname w/o RAG &  42.8          & 36.1     & 50.8                & 52.0      & 45.0        \\ \hline
LLaMA-2-13B w/o RAG                 & 52.8       & 44.1 & 62.6            & 61.1  & 54.8    \\
+ \modelname w/o RAG & 52.5           & 43.7     &  62.1               & 60.9      & 54.3        \\ \toprule
\end{tabular}}
\caption{Performance on MMLU in the setting without retrieval-augmented generation.}
\label{mmlu}
\end{table}

\end{document}